\documentclass[letterpaper, 10 pt, conference]{ieeeconf}  

\IEEEoverridecommandlockouts                              

\title{\LARGE \bf
LoopSR: Looping Sim-and-Real for Lifelong Policy Adaptation of Legged Robots
}

\author{Peilin Wu$^{1}$, Weiji Xie$^{1}$, Jiahang Cao$^{1}$, Hang Lai$^{1}$, and Weinan Zhang$^{1}$
\thanks{$^{1}$Dept. of Computer Sci. and Eng., Shanghai Jiao Tong University, China}
}

\usepackage{amsthm, amsfonts}
\usepackage{amsmath}
\usepackage{amssymb}
\usepackage{multirow}
\usepackage{multicol}
\usepackage{stfloats}
\usepackage{algorithm}
\usepackage{algorithmicx}
\usepackage[noend]{algpseudocode}
\usepackage{booktabs}
\usepackage{float}
\usepackage{graphicx}
\usepackage{array}
\usepackage{tabularx}
\usepackage[square,sort,comma,numbers,compress]{natbib}
\usepackage{hyperref}
\usepackage{colortbl}
\usepackage{nicefrac}
\usepackage{bm}
\usepackage[T1]{fontenc}
\usepackage{makecell}

\usepackage{lineno}
\usepackage{color}
\usepackage{xcolor}

\begin{document}

\maketitle
\thispagestyle{empty}
\pagestyle{empty}

\begin{abstract}

Reinforcement Learning (RL) has shown its remarkable and generalizable capability in legged locomotion through sim-to-real transfer. However, while adaptive methods like domain randomization are expected to enhance policy robustness across diverse environments, they potentially compromise the policy's performance in any specific environment, leading to suboptimal real-world deployment due to the No Free Lunch theorem. To address this, we propose LoopSR, a lifelong policy adaptation framework that continuously refines RL policies in the post-deployment stage. LoopSR employs a transformer-based encoder to map real-world trajectories into a latent space and reconstruct a digital twin of the real world for further improvement. Autoencoder architecture and contrastive learning methods are adopted to enhance feature extraction of real-world dynamics. Simulation parameters for continual training are derived by combining predicted values from the decoder with retrieved parameters from a pre-collected simulation trajectory dataset. By leveraging simulated continual training, LoopSR achieves superior data efficiency compared with strong baselines, yielding eminent performance with limited data in both sim-to-sim and sim-to-real experiments. Please refer to \hyperref[website]{https://peilinwu.site/looping-sim-and-real.github.io/} for videos and code.

\end{abstract}


\section{Introduction}


Reinforcement Learning (RL) has seen significant progress as a viable alternative for robotic control, but applying it to real-world scenarios remains challenging. Advances in simulation systems \cite{IsaacGym, mujoco} and extensive parallel training \cite{rudin2022massive} have enhanced policy performance and robustness, but the sim-to-real gap persists.

Adaptive methods such as domain randomization \cite{lee2020challenging, tan2018sim2real, randomization1, randomization2, randomization3} are widely used to address the gap and improve robustness. However, given RL methods' inherent requirement of direct interaction with the environment, and along with No Free Lunch Theorem \cite{NFL_optim, NFL_supervised}, which suggests a potential trade-off between generalization and specific performance, purely training in the simulation with inaccurate state occupancy distribution will limit the policy's performance beforehand. 

Leveraging the data in the real world where the robot is directly situated is an intuitive solution. Nevertheless, it faces obstacles. Firstly, collecting real-world data is notoriously expensive while RL methods are extremely data-hungry. An applicable policy generally requires several months of real-world experience which is hardly affordable. Secondly, absent privileged knowledge (heightfield, contact force, mass, friction, etc) in real-world settings complicates tasks on tricky terrains. One example is the stairs, where a blind robot with only proprioceptive observations  (i.e. joint position \& velocity) needs extensive exploration to learn to lift its legs, while the one with precise height information can better find such a solution. On top of that, noisy observation also leads to training instability.

Previous works have attempted to cope with these impediments by reshaping reward functions \cite{smith2022walk, smith2022legged}, taking advantage of sample-efficient algorithms \cite{smith2022walk, smith2022legged, smith2024grow}, or harnessing model-based RL \cite{wu2023daydreamer}. However, these methods, trained directly in the real world, fail to produce superior locomotion policies and are vulnerable during training compared to zero-shot methods.


\begin{figure}
    \centering
    \includegraphics[width=\linewidth]{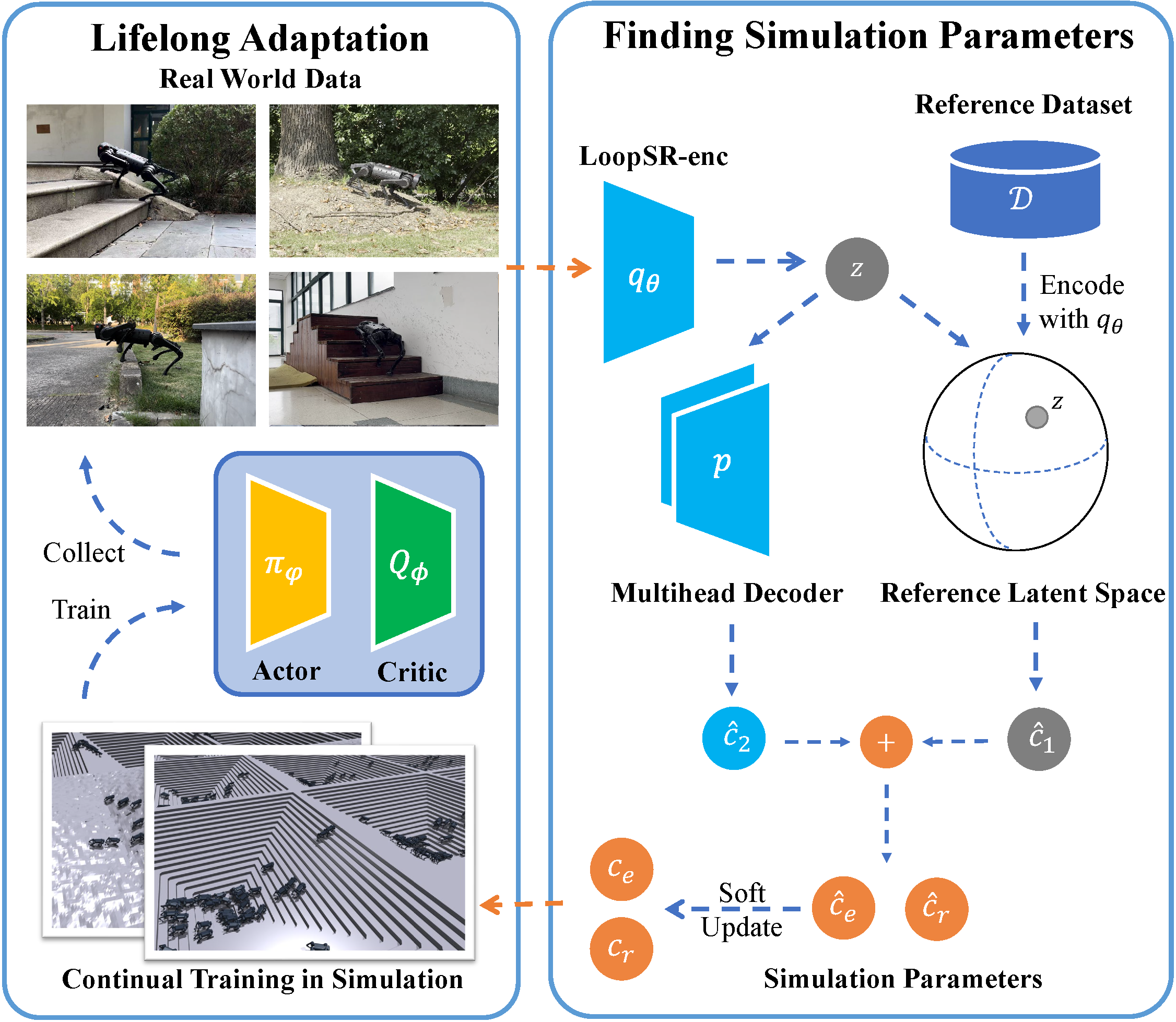}
    \vspace{-0.65cm}
    \caption{Overview of our proposed LoopSR.}
    \label{fig:overview}
    \vspace{-0.65cm}
\end{figure}
 
Drawing inspiration from how animals use familiar environments to navigate new ones, we investigate to bridge the way back to the controllable simulation environment rather than trapping ourselves in the capricious and noisy real world. This approach offers all-around benefits. In one way, we can sidestep the problem of redesigning a delicate reward function corresponding to the limited observation, and take full advantage of a well-polished simulation training process \cite{IsaacGym, rudin2022massive, lee2020challenging, rma}. Additionally, it requires minimal real-world data while still living off the data abundance available in the simulation.

 
In this work, we present LoopSR, an effective pipeline for lifelong policy adaptation in legged robots depicted in Fig.~\ref{fig:overview}. At its core, LoopSR utilizes a transformer-based encoder \cite{transformer} to leverage the latent space. By adopting an autoencoder architecture \cite{TAE, VAE} and contrastive loss \cite{simsaim, infoNCE, con_gcrl, CURL}, the model extracts essential features for reconstructing simulation dynamics, ensuring both effectiveness and generalizability. 

We implement learning-based and retrieval-based methods to derive simulation parameters from the latent variable, which compensate for each other in accuracy and robustness to out-of-distribution situations. Through comprehensive evaluations, LoopSR outperforms strong baselines in both the IsaacGym simulated environment \cite{IsaacGym} and the real-world scene.

In a nutshell, the contributions of this paper can be summarized as follows:
\begin{itemize}
    \item We proposed a novel lifelong policy adaptation method LoopSR that first incorporates real-world policy adaptation into simulated policy pretraining, achieving constant improvement during real-world deployments. 
    \item We propose a representation-learning approach and construct an all-encompassing latent space for unlabeled trajectories. With the innovative combination of learning-based and retrieval-based methods, we successfully identify and rebuild the real world in simulation.
    \item We empirically validate LoopSR in both simulation and real-world environments. The results demonstrate the prominence of LoopSR compared with previous works.
\end{itemize}

\section{Related Works}
\subsection{Representation Learning in RL}
Representation learning \cite{VAE, representation_survey,repre_multimodal_sur, repre_con_sur} has made up the bedrock for modern deep learning algorithms, and the field of RL and control is no exception. Previous works have made versatile attempts to take advantage of the representation learning and latent space \cite{repre_rl2, Hilbert, repre_rl}, including contrastive learning \cite{CURL, R3M}, reward function encoding 
\cite{reward_encoding, reward_repre_gcrl}, domain/task-specific representation \cite{TAE, robot_specific, skill-transfer}, and gait encodings \cite{walk_these_ways, Gait_representation}, boosting multi-modal input \cite{repre_multimodal_sur} or transferring through different tasks and embodiment \cite{robot_specific}. 

Typically, the domain representation is widely utilized for the task of meta-RL and transfer learning 
\cite{transfer_learning2, transfer_learning_survey}. However, real-world scenarios present a challenge as goal reward functions are not readily available, where we are obliged to focus on learning the representation of the transitions themselves rather than relying on labeled tasks. 
 
\subsection{Learning in the Real World}
Continual learning in RL \cite{CRL_definition} is a daunting task as the result of the compounding error in drifted distribution. Works have been done to address the problem caused by protean data distribution through experience replay \cite{CRL_SER, CRL_replay}, distillation \cite{CRL_PR_distill}, and reusing parts of model \cite{CRL_DET}, etc. However, when it comes to real-world learning, the situation becomes progressively difficult due to noisy input, limited observation, and inaccessible reward functions. 

\cite{smith2022walk, smith2022legged, smith2024grow} are among the few works tapping into this tricky problem. Instead of sim-to-real transfer, these works have made attempts to completely learn the policy in the real world by introducing reshaped reward functions and techniques of restricting action space to boost effective training.
However, these algorithm needs careful designing in action space constraints and rewards to reluctantly obtain effective training and executable gaits, while failing to outstrip zero-shot methods.


With world models exhibiting competence in heterogeneous downstream tasks 
\cite{dreamer}, \cite{wu2023daydreamer} employ the Dreamer world model in the real-world learning task. The model-based method, though a practical alternative, still suffers from the problem of deficient privileged knowledge. 
The model's accuracy in featuring real-world information is also worth questioning in comparison with the simulation, as model-based methods are susceptible to pervasive model biases. The vulnerability of training also places the risk that convergence is not guaranteed facing the composition of model biases and compounding error. 

\section{Methodology}
As shown in Fig. \ref{fig:overview}, our proposed method LoopSR (\textbf{Loop}ing \textbf{S}im and \textbf{R}eal) supports lifelong policy adaptation in quadrupedal locomotion. In the following subsections, we first introduce the preliminaries (Sec. \ref{sec:preliminaries}) and provide an overview (Sec. \ref{sec:overall}) of our algorithm. Then we illustrate our core training framework in Sec. \ref{sec:TrajEnc} and provide important technical details in Sec. \ref{sec:implement-details}. Finally, we present a script of theoretical analysis to support the validity (Sec. \ref{sec:analysis}).
\subsection{Preliminaries}
\label{sec:preliminaries}
We formulate the training process in the simulator as a Partially Observable Markov Decision Process (POMDP). We define the MDP in the simulation as $\mathcal{M}=(\mathcal{S}, \mathcal{O}, \mathcal{A}, P, r, \mathcal{\gamma})$, where $\mathcal{S}, \mathcal{O}, \mathcal{A}$ denotes the state space, observation space, and action space respectively. 
$P(s_{t+1}|s_t,a_t)$ denotes the transition function, and the reward function is denoted as $r(s, a)$. 
$\gamma$ denotes the discount factor. We define the privileged environment parameters $c \in \mathcal{S} \backslash \mathcal{O}$ used to construct the simulation environment. Specifically, 
$c = [c_e, c_r]$ contains the knowledge of the terrain distribution $c_e$ (terrains are divided into 5 categories for feasibility) and robot-concerned parameters $c_r$ (mass, motor strength, etc.). The goal of the RL method is to get the policy to maximize the expected discounted return:
\begin{equation}
\pi^*:= \arg\max_{\pi}\mathbb{E}_{a_t \sim \pi(\cdot|o_t)}\Big[\sum\nolimits_{t=0}^{\infty}\gamma^t r(s_t,a_t)\Big].
\end{equation}


In the task of the lifelong policy adaptation, we need to adapt the pretrained policy to optimal in the real scene with the POMDP $\mathcal{M}^R=(\mathcal{S}^R, \mathcal{O}^R, \mathcal{A}^R, P^R, r^R, \mathcal{\gamma}^R)$, where $r^R$ is implicit without access. 
In the real world, we can collect trajectories constantly but at an extremely low speed, at least $1000$ times slower than in the simulation. 
In the later analysis, we take $\mathcal{M}$ and $\mathcal{M}^R$ to represent the training environment and the testing one.

\subsection{Looping Sim and Real}
\label{sec:overall}
Fig.~\ref{fig:overview} and pseudocode~\ref{alg} outline LoopSR’s two-stage pipeline: pretraining and lifelong adaptation. During pretraining, LoopSR learns a deployable policy in the simulation while constructing an informative latent trajectory space by training an encoder-decoder network. In the adaptation stage, it encodes real-world trajectories into latent variables and refines simulation dynamics by leveraging stored rollouts and learned representations.

For pretraining, we employ Proximal Policy Optimization (PPO) \cite{PPO} in the IsaacGym environment following \cite{rudin2022massive}. The policy is based on DreamWaQ \cite{dreamwaq} and AMP \cite{AMP1, amp2}, utilizing an Asymmetric Actor-Critic architecture with expert motion data and domain randomizations \cite{randomization1, randomization2, randomization3} to enable real-world deployment. Meanwhile, LoopSR constructs a latent trajectory space by training the trajectory encoder (LoopSR-enc) alongside a multi-head decoder using collected rollouts and privileged information. These rollouts are stored as a reference dataset for later adaptation.

A collected real-world trajectory is encoded into a latent variable $z^R$ during the adaptation stage. The estimated $\hat{c}$ is then estimated by combining retrieval from stored rollouts ($\hat{c}_{1}$) and the decoded model output ($\hat{c}_{2}$). The final $\hat{c}$ configures the IsaacGym simulation.

\begin{algorithm}[!ht]
  \caption{The pipeline of \textbf{LoopSR}}
  \label{alg}
  \begin{algorithmic}[1]
    \Require pretraining iterations $n$, fusion ratio $\alpha$, soft update ratio $\tau$
    \State Initialize policy $\pi$, LoopSR-enc $q_{\theta}$, Decoders $q_{\psi}, p_e, p_r$ and dataset $\mathcal{D} \leftarrow \emptyset$
    \State Set $c^{curr} =[c^{curr}_e, c^{curr}_r]$ as the average parameter
    ~\\
    \State // \textit{Pretrain policy in simulation}
    \For{$i=1$ to $n$}
    \State Rollout and store the trajectories, relevant terrain distribution, and robot parameter $(\mathcal{T}, c_e, c_r) \rightarrow \mathcal{D}$  
    \State Update $\pi$ through PPO Loss
    \State Update $q_{\theta}, q_{\psi}, p_e, p_r$ through Loss in (\ref{loss_vae}), (\ref{loss_con}) and (\ref{loss_head})
    \EndFor
    \State Get reference dataset 
    
    $\mathcal{D}_z \leftarrow \{(z, c_e, c_r) \big| z = q_{\theta}(\mathcal{T}), \forall \mathcal{T}, c_e, c_r \in \mathcal{D} \}$
    ~\\
    \State // \textit{Lifelong policy adaptation}
    \Repeat
      \State Collect testing environment trajectory $\mathcal{T}$ with $\pi$
      \State Get $ z = q_{\theta}(\mathcal{T}) $ 
      \State Retrieve $N$ nearest neighbours $z_1, z_2, \cdots, z_N$ of $z$ in $\mathcal{D}_z$, and average the corresponding parameter $c_1, c_2, \cdots, c_N$ to get  $\hat{c}_{1} = [c_e', c_r']$
      \State Get $\hat{c}_{2} = [p_e(z), p_r(z)]$
      \State Fuse predicted parameters $\hat{c} = \alpha \hat{c}_{1} + (1 - \alpha) \hat{c}_{2}$
      \State Soft update $c^{curr} = \tau c^{curr} + (1 - \tau) \hat{c} $
      \State Build the simulation according to $c^{curr}$ in simulation and continue training $\pi$
    \Until{\textbf{forever}}
  \end{algorithmic}
\end{algorithm}
\vspace{-0.45cm}

\subsection{Network Architecture}
\label{sec:TrajEnc}
\begin{figure*}[!ht]
  \centering
  \vspace{0.15cm}
  \includegraphics[width=0.95\textwidth]{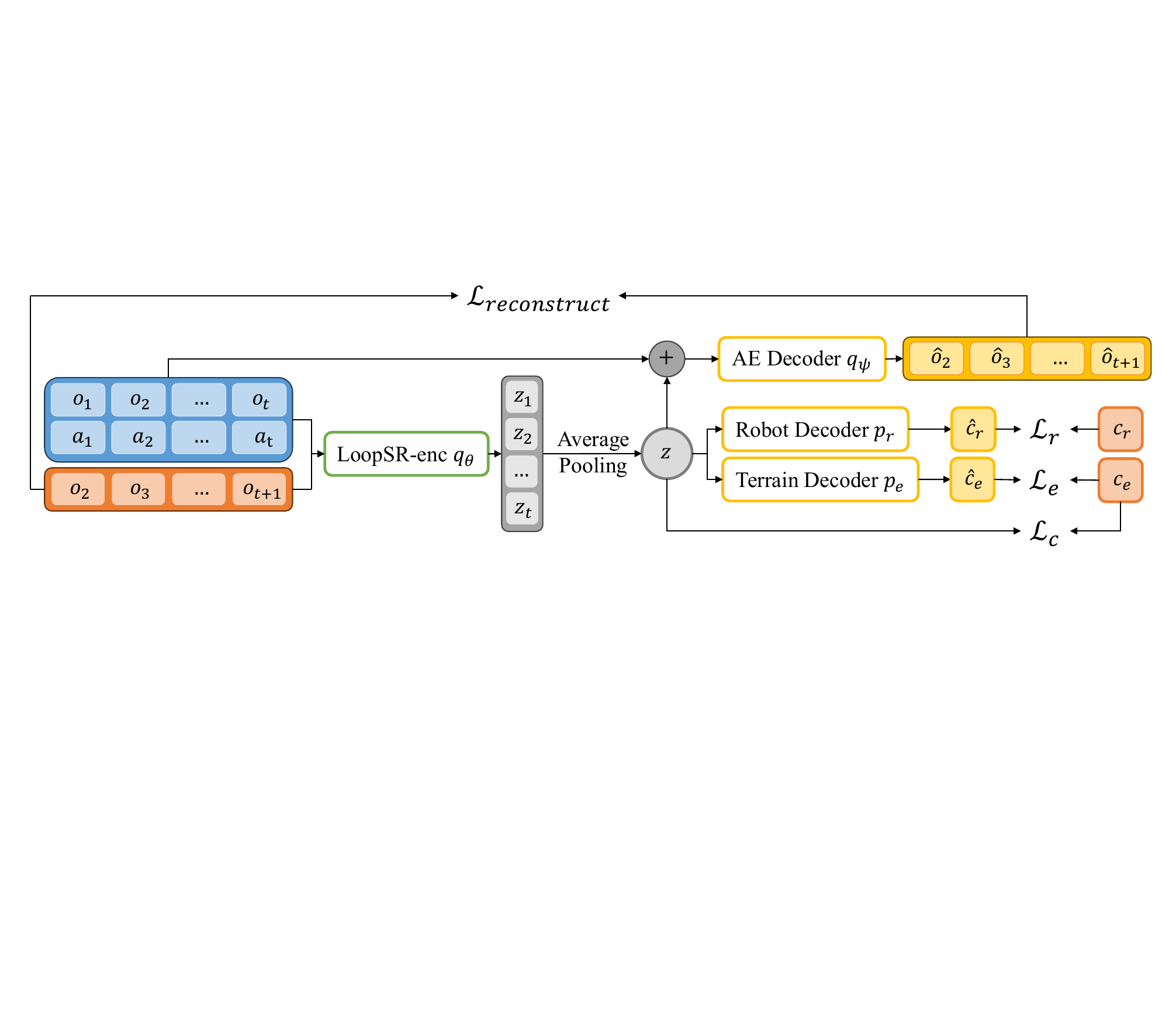}
  \vspace{-0.25cm}
  \caption{Illustration of the core network architecture. The whole trajectory is input to a transformer backbone similar to \cite{DT}, with output $z$ derived from average pooling. The decoders are separately designed to disentangle different traits from the trajectory.}
  \label{fig:framework}
  \vspace{-0.5cm}
\end{figure*}

To the end of retaining the corresponding parameters of the testing environment, we design the network to cash in the latent space for knowledge distillation from real-world trajectories inspired by the advancements in representation learning methods \cite{TAE, VAE, repre_seq}. 
The network architecture and its working flow are exhibited in Fig.~\ref{fig:framework}. 

\textbf{Transformer-based Encoder.} The key encoder, LoopSR-enc draws on Decision Transformer \cite{DT} or specifically follows \cite{GPT} as the backbone while substituting the embedding of returns-to-go with the one of the next observation. 
LoopSR-enc $q_{\theta}(z|o_{1:t}, a_{1:t}, o_{2:t+1})$ receives the total trajectory without reward notation $\mathcal{T} = (o_{1:t}, a_{1:t}, o_{2:t+1})$ as input. The output latent variable $z \in \mathbb{R}^{32}$ is the average pooling of the latent $z_i$ from every timestep $i$ and is modeled as a normal distribution following \cite{VAE}. To support efficient learning \cite{Gait_representation, ASE}, we project $z$ to an $l_2$ sphere to bound the latent space. 

\textbf{Autoencoder Architecture.} To ensure that $z$ captures enough features from the original trajectory, the training process is modeled as autoencoder(AE)-like with a decoder $q_{\psi}(\hat{o}_{n+1}|z, o_n, a_n)$ to reconstruct the new observations $o_{n+1}$ based on the latent variable $z$, the observations $o_n$ and the actions $a_n$ inspired by \cite{TAE}. We follow the common approach to use L2 distance as a supplant for the untractable log probability \cite{TAE, repre_seq, repre_sto_vvp} of the next state prediction:
\begin{equation}
\label{loss_vae}
    \mathcal{L} =\sum_{n=0}^{t}\big[ o_{n+1} - q_{\psi}\big(q_{\theta}(\mathcal{T}), o_n, a_n\big)\big]^2, \forall \mathcal{T} \in \mathcal{B}.
\end{equation}

\textbf{Contrastive Loss.} To guide the model to discriminate different terrains implicitly, we apply the contrastive loss in the form of supervised InfoNCE loss similar to \cite{supervised_con}. 
The loss is defined to maximize the inner product of $z$ from the trajectories with the same terrain type label, whose theoretical background of maximizing the mutual information between positive samples and the latent representation is well-discussed in \cite{infoNCE, supervised_con}.
\begin{equation}
\label{loss_con}
    \mathcal{L}_c = - \frac{1}{N} \sum_{i=1}^{N} \mathbf{1}_{c_e^i=c_e^j} \log \frac{\exp(z^i\cdot z^j)}{\sum_{j=1}^{N} \mathbf{1}_{i\neq j}\exp(z^i\cdot z^j)},
\end{equation}
where $N$ is the batch size and the superscript $i, j$ denotes the index of different trajectories.

\textbf{Multi-head Decoder.} We also design a multi-head decoder $p_e$ and $p_r$ to separately extract the terrain distribution and the robot proprioceptive parameters for disentanglement of different environment characteristics. We introduce $c_e$, $c_r$, and $\mathcal{R}$ to represent the environment terrain distribution, robot parameter, and the randomization range of parameters in the training process. The losses are designed as follows:
\begin{equation}
\label{loss_head}
    \mathcal{L}_{e} = D_{KL}\big(p_e(z)||c_e\big), \ \ \mathcal{L}_{r}= \frac{|p_r(z) - c_r|}{|\mathcal{R}|},
\end{equation}
where $|\mathcal{R}|$ denotes the interval length of the randomization range.
In Sec.~\ref{sec:ablation}, we conduct the ablation study to discuss the necessity of all loss terms.

It's noteworthy that $z$ learns the manifold of simulation environments as LoopSR-enc and decoders are trained only on simulation data. 
This entails LoopSR with the potential capacity to find the best set of simulation parameters that generate a similar trajectory in the testing environment.

\subsection{Implementation Details}
\label{sec:implement-details}

\textbf{Soft Update.} 
We implement the soft update of estimated environment parameters to avoid deleterious abrupt changes in terrain distribution and robot parameters. 
Specifically, the estimated environment parameters is updated through $c^{curr} = \tau c^{curr} + (1-\tau) \hat{c} $. $\tau$ is set to be $0.7$. 
Such a method has the adaptive ability to attune to the environment setting which is not in the initial reference dataset. The necessity of soft update is ablated in Sec.~\ref{sec:ablation}.

\textbf{Retrieval Process.} 
Regarding training safety, we additionally employ the retrieval method 
to preclude drastically mistaken results in the face of out-of-distribution (o.o.d.) instances from learning-based methods. The final result is weighted as $\hat{c} = \alpha \hat{c}_{1} + (1 - \alpha)\hat{c}_{2}$ with $\alpha$ set to be $0.8$. 

\subsection{Theoretical Analysis}
\label{sec:analysis}
We provide a theoretical analysis here to explain the effectiveness of LoopSR. With the No Free Lunch Theorem, we demonstrate that wide domain randomization during policy training leads to suboptimal strategy in a specific environment.
The influence on policy performance of the differences between the training environment and the testing one can be modeled as the discrepancy between optimal value functions. in the training environment and the testing one is derived in (\ref{eq-full}).
\begin{equation}
    \label{eq-full}
    \begin{aligned}
        &\:\:\:\:\: \big|Q^*(s,a)-Q^*_R(s,a)\big|\\
        &=\Delta r + \underbrace{\gamma\sum_{s',a'}P(s'|s,a)\big[\pi^*Q^*- \pi_R^*Q^*_R \big]}_{(\romannumeral 1) \text{~total policy discrepancy}} \\
        &\:\:\:\:\:+\underbrace{\gamma\sum_{s',a'}P(s'|s,a)\pi_R^*Q^*_R -\gamma\sum_{s'',a''}P^R(s''|s,a)\pi_R^*Q^*_R}_{(\romannumeral 2) \text{~transition discrepancy}} \\
        &\leq \frac{\Delta r}{1 - \gamma} + \frac{2\gamma r_{max}}{(1 - \gamma)^2} \epsilon_{\pi} + \frac{\gamma}{1-\gamma} V_R^*(s') \epsilon_{\rho},
    \end{aligned}
\end{equation}
where 
\begin{equation*}
    \Delta r = r - r^R,
\end{equation*}
\begin{equation*}
    \epsilon_{\pi} = \sup_s D_{TV}(\pi^*(\cdot|o)||\pi^*_R(\cdot|o)),
\end{equation*}
\begin{equation*}
    \epsilon_{\rho} = \mathbb{E}_s [D_{TV}(\rho(s)||\rho^R(s))].
\end{equation*}

In (\ref{eq-full}), $Q^*$ and $Q^*_R$ are optimal value functions in $\mathcal{M}$ and $\mathcal{M}^R$ respectively, and $V_R^*(s')$ as the optimal state value function on $\mathcal{M}^R$. $\Delta r$ informs the discrepancy in reward functions, which is negligible for the  substantiated validity of $r$ from previous works \cite{rudin2022massive, lee2020challenging, egocentric, rma}. $\epsilon_{\pi}$ models the discrepancy caused by policy differences, which should approach $0$ with sufficient model capacity as $\pi^*$ and $\pi^*_R$ share the same input with detailed reasoning in \cite{HIB}. 

The third term $\epsilon_{\rho}$ denotes the discrepancy induced by the shift in state occupancy distribution, where $\rho(s)$ is closely related to the innate state occupancy distribution (e.g. the state distribution of the stair terrain differs from the plain one). However, wide domain randomization inherently leads to large $\epsilon_{\rho}$ with larger environment diversity. On the other hand, LoopSR introduces the process of building a digital twin of the real world and gradually training the policy on similar environments, thus efficiently mitigating such an error and optimizing the policy performance. 

\section{Experiments}
\begin{table*}[!ht]
    \centering
    \vspace{0.2cm}
    \caption{The Result on Different Terrains in IsaacGym Simulation}
    \label{table:sim-terrain}
    \begin{tabular*}{\textwidth}{l|c|c|c|cccc}
        \toprule
        \textbf{Terrain} & 
        \textbf{Difficulty} & 
        \textbf{Expert} & 
        \textbf{Origin (DreamWaQ)} & 
        \textbf{LoopSR (Ours)} & 
        \textbf{RMA} & 
        \textbf{SAC} & 
        \textbf{World Model} \\ 
        \midrule
        \multirow{3}{*}{\textit{Stairs}} & $0.122$ & $1.3731\scriptstyle{\pm0.0002}$ & $1.2238\scriptstyle{\pm0.00189}$ & \cellcolor[rgb]{0.9,1.0,0.9}\bm{$1.3084\scriptstyle{\pm0.00062}$} & 
        $1.2183\scriptstyle{\pm0.02329}$ &
        $0.2662\scriptstyle{\pm0.07617}$ & $0.1274\scriptstyle{\pm0.03480}$
         \\
        &$0.176$ &$1.3519\scriptstyle{\pm0.0003}$ &  $1.1874\scriptstyle{\pm0.00043}$ & \cellcolor[rgb]{0.9,1.0,0.9}\bm{$1.2522\scriptstyle{\pm0.00023}$} & 
        $1.1476\scriptstyle{\pm0.08996}$ &
        $0.0386\scriptstyle{\pm0.00506}$ & 
        $0.0619\scriptstyle{\pm0.01521}$ 
        \\
        &$0.23$ & $1.2414\scriptstyle{\pm0.0354}$& $0.7553\scriptstyle{\pm0.12505}$ & \cellcolor[rgb]{0.9,1.0,0.9}\bm{$1.1330\scriptstyle{\pm0.00813}$} &  
        $0.3680\scriptstyle{\pm0.19475}$ &
        $0.0253\scriptstyle{\pm0.00024}$ & 
        $0.0436\scriptstyle{\pm0.02800}$ 
        \\
        \midrule
        \multirow{3}{*}{\textit{Slope}} & $0.16$ &  $1.4025\scriptstyle{\pm0.00001}$ & $1.3683\scriptstyle{\pm0.00003}$ & \cellcolor[rgb]{0.9,1.0,0.9}\bm{$1.3908\scriptstyle{\pm0.00001}$} & 
        $1.3048\scriptstyle{\pm0.00038}$ &
        $0.1877\scriptstyle{\pm0.04915}$ &
        $0.2813\scriptstyle{\pm0.01686}$ 
        \\
        &$0.4$ & $1.3478\scriptstyle{\pm0.00012}$ & $1.1968\scriptstyle{\pm0.00087}$ & \cellcolor[rgb]{0.9,1.0,0.9}\bm{$1.3417\scriptstyle{\pm0.00020}$} & 
        $1.1707\scriptstyle{\pm0.00694}$ &
        $0.1539\scriptstyle{\pm0.05367}$ &
        $0.2222\scriptstyle{\pm0.02665}$ 
        \\
        &$0.6$ & $1.1758\scriptstyle{\pm0.01412}$ & $0.8384\scriptstyle{\pm0.11217}$ &  \cellcolor[rgb]{0.9,1.0,0.9}\bm{$1.0746\scriptstyle{\pm0.03184}$} & 
        $0.9429\scriptstyle{\pm 0.00661}$&
        $0.0622\scriptstyle{\pm 0.02096}$ &
        $0.1227\scriptstyle{\pm0.03877}$ 
        \\
        \midrule
        \multirow{3}{*}{\textit{Discrete}} & $0.13$ & $1.3519\scriptstyle{\pm0.00137}$ & $1.2202\scriptstyle{\pm0.01363}$ & \cellcolor[rgb]{0.9,1.0,0.9}\bm{$1.2648\scriptstyle{\pm0.00193}$} & 
        $0.9072\scriptstyle{\pm0.01461}$&
        $0.3354\scriptstyle{\pm0.05142}$&
        $0.1607\scriptstyle{\pm0.04661}$
        \\
        &$0.19$ & $1.1638\scriptstyle{\pm0.14744}$ & $0.9932\scriptstyle{\pm0.06514}$ & \cellcolor[rgb]{0.9,1.0,0.9}\bm{$1.0918\scriptstyle{\pm0.01214}$} &  
        $0.7200\scriptstyle{\pm 0.09696}$&
        $0.0354\scriptstyle{\pm 0.05214}$&
        $0.1280\scriptstyle{\pm 0.04706}$ 
        \\
        &$0.25$ & $0.8681\scriptstyle{\pm0.26708}$ & $0.4056\scriptstyle{\pm0.21618}$ & \cellcolor[rgb]{0.9,1.0,0.9}\bm{$0.8379\scriptstyle{\pm0.19660}$} & 
        $0.1744\scriptstyle{\pm0.04831}$&
        $0.0138\scriptstyle{\pm0.04024}$& $0.1768\scriptstyle{\pm0.01260}$
        \\
        \bottomrule
        \end{tabular*}
        \vspace{-0.5cm}
\end{table*}




\subsection{Sim-to-Sim Experiment}
\label{sec:sim-results}
\textbf{Overview.} Although simulation settings cannot perfectly replicate real-world complexities, they offer a controlled environment for conducting statistical experiments, facilitating comparisons when direct real-world computation is impractical. We leverage IsaacGym \cite{IsaacGym} for training and simulation, 
and the training code is developed based on \cite{rudin2022massive}. 

Referred to Sec.~\ref{sec:overall}, the experiment consists of pertaining and adaptation stages. During pretraining, 4,096 robots are simulated across five distinct terrains (upward/downward slopes/stairs and plain), training in parallel for 30,000 iterations, with 24 simulation steps per iteration. 
During adaptation, data is collected in a testing environment chosen from three terrains, each with three difficulty levels (details see Table~\ref{table:sim-terrain}), where the environment with the highest difficulty is an out-of-distribution one. 
Privileged knowledge is not accessed in the testing environment to better simulate real-world conditions. 

In general, LoopSR runs $100$ sim-to-real loops for each testing environment. In each loop, we collect $1e3$ steps in the testing environment and train 500 iterations in the newly constructed training environment. In total, each policy is trained on $1e5$ steps of collected data, equivalent to half an hour in a real-world experiment. For a fair comparison, all baselines share the same total training iterations of 80,000.
During evaluations, robots are commanded to walk forward at a velocity of $v_x^{cmd}=1.0m/s$ to assess their maximum performance. The reward functions follow \cite{rudin2022massive} and are averaged across simulation steps.



\textbf{Baselines.} The baselines are defined as follows:
\begin{itemize}
    \item \textbf{Expert} is a reference policy completely trained on the desired testing environment. The architecture follows DreamWaQ \cite{dreamwaq}.
    \item \textbf{Origin (DreamWaQ)} is a reference policy without refinement, representing a zero-shot sim-to-real workflow following DreamWaQ \cite{dreamwaq}.
    \item \textbf{RMA} is another zero-shot sim-to-real workflow following RMA \cite{rma}. Here, the teacher and student policy are trained for 80,000 iterations respectively.
    \item \textbf{SAC} follows \cite{smith2022walk} with SAC used in pretraining and continual training. 
    \item \textbf{World Model} indicates a model-based method following \cite{wu2023daydreamer} with Dreamer \cite{dreamer} as the backbone. 
\end{itemize}

\begin{figure}[!ht]
    \centering
    \vspace{-0.1cm}
    \includegraphics[width=0.9\linewidth]{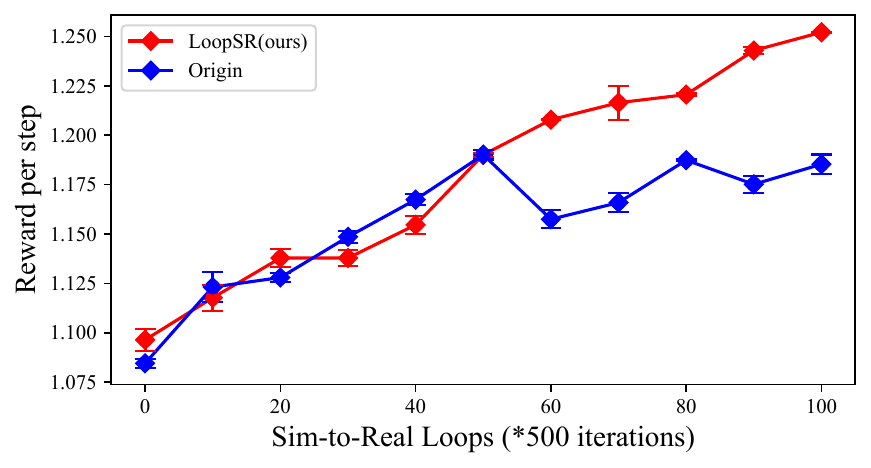}
    \vspace{-0.45cm}
    \caption{Continual training curve with \textbf{Origin (DreamWaQ)} baseline.
    }
    \label{fig:sr-loop}
    \vspace{-0.4cm}
\end{figure}

\textbf{Results.} The simulation results are shown in Tab.~\ref{table:sim-terrain}, where \textbf{LoopSR} achieves prominent enhancement and closely matches the performance of \textbf{Expert}, especially in challenging terrains like \textit{Stairs} and \textit{Discrete}. 
This outperformance over zero-shot sim-to-real methods (\textbf{Origin}, \textbf{RMA}) confirms LoopSR's success in continuous optimization. At the same time, direct real-world training approaches (\textbf{SAC}, \textbf{World Model}) mainly fail, typically owning to their fragility to the shift in state occupancy distribution. The reward curve concerning the sim-to-real loops is shown in Fig.~\ref{fig:sr-loop}, where each sim-to-real loop integrates simulated training of $500$ iterations. We find out that \textbf{LoopSR} surpasses the performance bottleneck which troubles the original sim-to-real baseline, demonstrating the necessity of policy adaptation.


\subsection{Sim-to-Real Evaluation}
\begin{figure}
    \centering
    \includegraphics[width=0.85\linewidth]{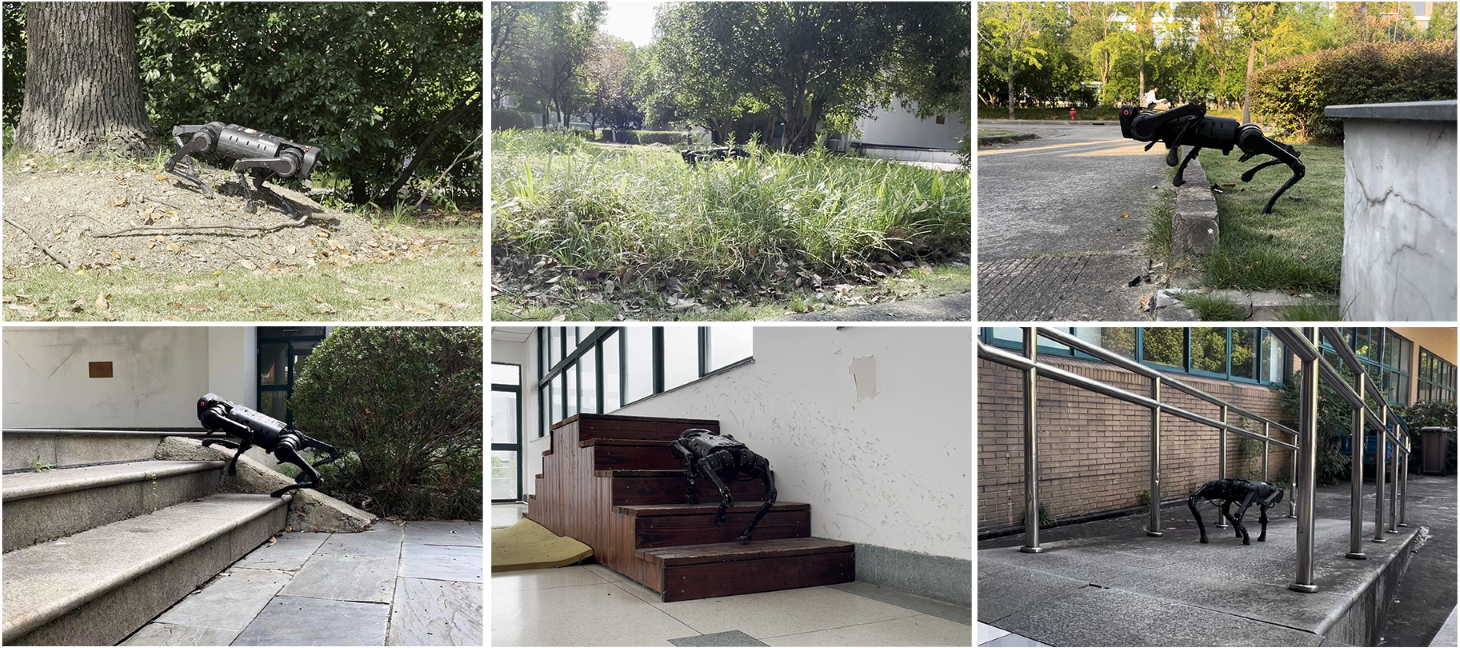}
    \vspace{-0.25cm}
    \caption{Visualization of real-world application scenes.}
    \label{fig:vis-real}
    \vspace{-0.65cm}
\end{figure}

\textbf{Overview.} For real-world evaluations, we utilize the Unitree A1 robot \cite{unitree} to facilitate real-world deployment. Real-world observation of A1 only includes proprioception $o_t \in \mathbb{R}^{45}$ without visual input. 
The action $a_t \in \mathbb{R}^{12}$ specifies target joint positions converted to joint torques via a PD controller. The testing environment consists of diverse real-world terrains, as depicted in Fig.~\ref{fig:vis-real}. 

The pretraining phase is identical to the sim-to-sim ones. For the policy adaptation phase, data collected in the real world are transmitted to a host computer with a GTX 4090 GPU to execute the continual training. To be more specific, we run 30 loops in total, and in each loop, we collect a batch of $5$ trajectories, each with length of $200$ timesteps (equal to $4$ seconds). After each batch of trajectories is collected, the agent is trained for $200$ iterations in the simulation. To simplify the data collection process, here we update the onboard policy every $10$ loops (equals $2000$ simulation iterations), which we find has little impact on the adaptive performance.


\begin{table}[!ht]
    \centering
    \caption{Time Cost to Traverse Different Terrains $\downarrow$}
    \begin{tabularx}{\linewidth}{ 
           >{\centering\arraybackslash}X 
           >{\centering\arraybackslash}X 
           >{\centering\arraybackslash}X 
           >{\centering\arraybackslash}X }
        \toprule
        \textbf{Domain} & \textbf{LoopSR} & \textbf{Origin} & \textbf{RMA} \\
        \midrule
        \textbf{Stair Up} & \cellcolor[rgb]{0.9,1.0,0.9}\bm{$4.33 \scriptstyle{\pm 0.11}$} & $5.01 \scriptstyle{\pm 0.18}$ & $6.97 \scriptstyle{\pm 0.17}$ \\
        \textbf{Stair Down} & \cellcolor[rgb]{0.9,1.0,0.9}\bm{$5.14 \scriptstyle{\pm 0.31}$} & $5.92 \scriptstyle{\pm 0.25}$ & $5.68 \scriptstyle{\pm 0.28}$ \\
        \textbf{Grass} & \cellcolor[rgb]{0.9,1.0,0.9}\bm{$3.80 \scriptstyle{\pm 0.12}$} & $4.04 \scriptstyle{\pm 0.23}$ & $5.54 \scriptstyle{\pm 0.27}$ \\
        \textbf{Slope Down} & \cellcolor[rgb]{0.9,1.0,0.9}\bm{$10.60 \scriptstyle{\pm 0.13}$} & $10.82 \scriptstyle{\pm 0.09}$ & $10.99 \scriptstyle{\pm 0.13}$ \\
        \textbf{Slope Up} & \cellcolor[rgb]{0.9,1.0,0.9}\bm{$11.18 \scriptstyle{\pm 0.13}$} & $11.39 \scriptstyle{\pm 0.14}$ & $11.59 \scriptstyle{\pm 0.09 }$ \\
        \bottomrule
    \end{tabularx}
    \vspace{-0.2cm}
    \label{table:real-time}
\end{table}

We compare LoopSR to Origin (DreamWaQ) and RMA baselines 
to highlight the improvement. Each evaluation includes $10$ trials to reduce stochasticity while obviating damage to robot hardware. The performance metrics include the time required to traverse a specified path and the success rate in demanding terrains.

\begin{table}[!ht]
    \centering
    \caption{The Success Rate in Challenging Terrains}
    \begin{tabularx}{\linewidth}{ 
           >{\centering\arraybackslash}X 
           >{\centering\arraybackslash}X 
           >{\centering\arraybackslash}X 
           >{\centering\arraybackslash}X }
        \toprule
        & \textbf{Stair Up} & \textbf{Stair Down} & \textbf{Pit}  \\
        \midrule
        \rowcolor[rgb]{0.9,1.0,0.9} \textbf{LoopSR} & \bm{$100$} & \bm{$100$} & \bm{$90$}   \\
        \textbf{Origin}  & $70$ & $90$ & $80$  \\
        \textbf{RMA} & $60$ & $80$ & $60$ \\
        \bottomrule
    \end{tabularx}
    \vspace{-0.2cm}
    \label{table:success_rate}
\end{table}
  
\textbf{Results.} The results in Tab.~\ref{table:real-time} and Tab.~\ref{table:success_rate} demonstrate that \textbf{LoopSR} achieves superior performance across all tasks compared with zero-shot sim-to-real methods \textbf{Origin} and \textbf{RMA}, and ensures better safety during operation. 
The results indicate that \textbf{LoopSR} enhances locomotion policy capacity and fulfills the purpose of lifelong policy adaptation.

\subsection{Empirical Study on Gaits}
To further illustrate the effect of our method, we take the stair terrain as an example and analyze the gaits.

\begin{figure}[!ht]
    \centering
    \vspace{-0.1cm}
    \includegraphics[width=0.9\linewidth]{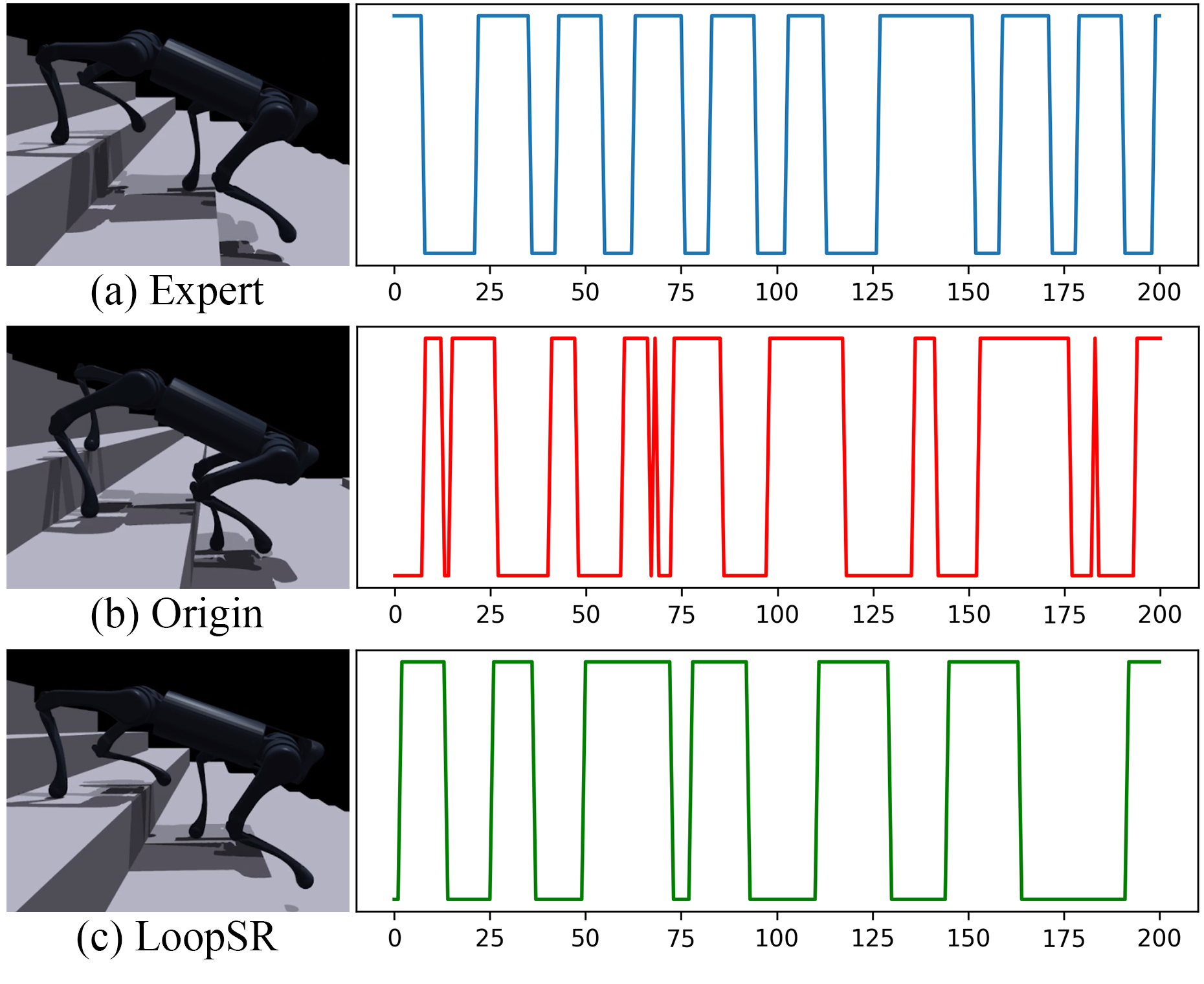}
    \vspace{-0.25cm}
    \caption{Visualization of simulation gaits and contact phase.}
    \label{fig:vis-sim}
    \vspace{-0.25cm}
\end{figure}

In the simulation, we visualize the robot's contact phase of one foot in Fig.~\ref{fig:vis-sim}, showing that our method gradually reaches the periodical gait of the expert. At the same time, the baseline without refinement frequently missteps.

\begin{figure}
    \centering
    \vspace{-0.5cm}
    \includegraphics[width=0.9\linewidth]{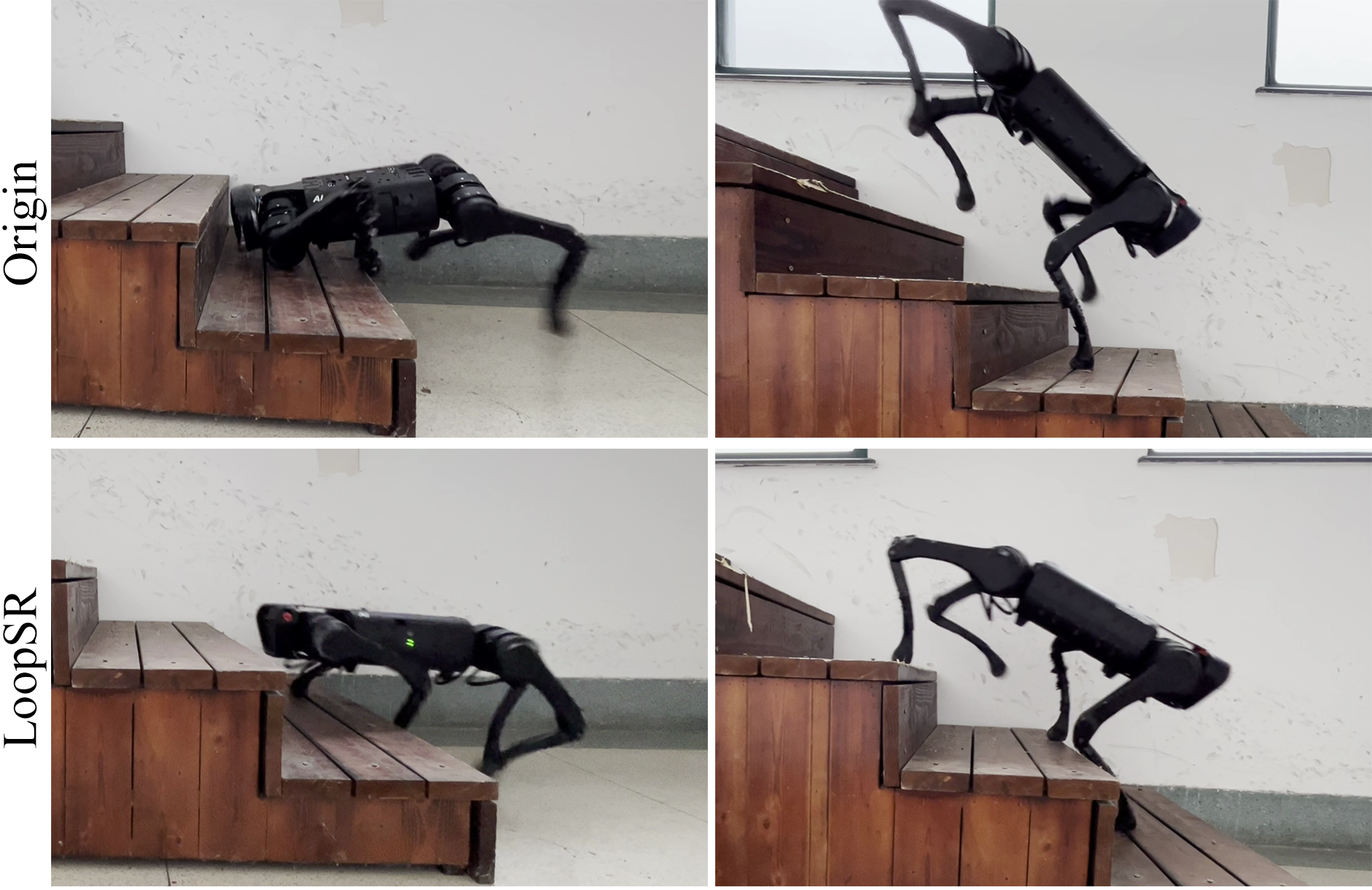}
    \vspace{-0.3cm}
    \caption{Comparison of motions from baseline \textbf{Origin} and \textbf{LoopSR}.}
    \label{fig:failures}
\end{figure}
In real-world experiments, 
We present snapshots of several jerky actions from Origin that triggered the safety module, along with corresponding gaits from the refined policy in Fig.~\ref{fig:failures}. The baseline reacts violently to terrain changes, producing unstable motions and raising the robot in the air, which risks damaging the hardware. In comparison, refined policy smoothly adapts to such variations, guaranteeing safety in real-world applications.

The dominance of our method in stair terrain proves that privileged knowledge, such as height fields and collisions, becomes crucial when facing challenging terrains, especially height-sensitive terrains.

\subsection{Ablation Study}
\label{sec:ablation}
We conduct the ablation study to verify each component of LoopSR, including each loss term and the soft update technique. 
The variants are designed as:
\begin{itemize}
    \item \textbf{LSR-w/o-con} eliminates the contrastive loss in the training process of LoopSR-enc.
    \item \textbf{LSR-w/o-AE} dismisses the reconstruction loss in the training process of LoopSR-enc.
    \item \textbf{LSR-w/o-su} shares the same LoopSR-enc with \textbf{LSR} and excludes the soft update in the adaptation stage.
\end{itemize}


For \textbf{LSR-w/o-con} and \textbf{LSR-w/o-AE}, we evaluate the prediction accuracy for simulation parameters to test the variants' ability to reconstruct a more similar simulation. For \textbf{LSR-w/o-su}, we compare the final reward across different terrains to discuss how the strategy in building simulation counterparts matters.

From Tab.~\ref{table:ablation}, \textbf{LSR} with all losses outperforms variants in accurately predicting simulation parameters. 
The comparable failure of \textbf{LSR-w/o-con} signifies the importance of the contrastive loss \cite{amp2} in building an interpretable latent space. Besides, \textbf{LSR} also surpasses \textbf{LSR-w/o-AE} with a considerable margin, which proves that the autoencoder structure is beneficial in extracting enough information from the trajectories. Furthermore, as Tab.~\ref{table:ablation-su} shows, \textbf{LSR-w/o-su} drastically underperforms on demanding terrains, demonstrating the necessity of soft update.

\begin{table}[!ht]
    \centering
    \caption{Ablation of Loss Components}
    \begin{tabularx}{\linewidth}{ c | c | 
    >{\centering\arraybackslash}X 
    >{\centering\arraybackslash}X
    >{\centering\arraybackslash}X }
        \toprule
        \multicolumn{2}{c}{\textbf{Domain}} & \textbf{LSR} & \textbf{LSR-w/o-con} & \textbf{LSR-w/o-AE} \\
        \midrule
        \multirow{3}{*}{\textbf{\makecell{Terrain \\ Accuracy }$\uparrow$}}  
        & \textit{Stair} & \cellcolor[rgb]{0.9,1.0,0.9} \bm{$92.1$} & $66.5$ & $74.5$ \\
        & \textit{Slope} & \cellcolor[rgb]{0.9,1.0,0.9} \bm{$81.2$} & $48.7$ & $52.5$ \\
        & \textit{Plain} & \cellcolor[rgb]{0.9,1.0,0.9} \bm{$87.9$} & $81.6$ & $50.9$ \\
        \midrule
        \multirow{2}{*}{\textbf{\makecell{Parameter \\ MSE }$\downarrow$}}  
        & \textit{Friction} & \cellcolor[rgb]{0.9,1.0,0.9} \bm{$19.08$} & $22.72$ & $23.28$ \\
        & \textit{Mass} & \cellcolor[rgb]{0.9,1.0,0.9} \bm{$24.37$} & $24.88$ & $24.71$ \\
        \bottomrule
    \end{tabularx}
    \vspace{-0.4cm}
    \label{table:ablation}
\end{table}


\begin{table}[!ht]
    \centering
    \caption{Ablation of Soft Update}
    \begin{tabularx}{\linewidth}{ 
           >{\centering\arraybackslash}X |
           >{\centering\arraybackslash}X 
           >{\centering\arraybackslash}X }
        \toprule
        \textbf{Domain} & \textbf{LSR} & \textbf{LSR-w/o-su} \\
        \midrule
        \textbf{Stair} & \cellcolor[rgb]{0.9,1.0,0.9}\bm{$1.2522\scriptstyle{\pm0.00023}$} & $0.5084$ \\\textbf{Slope} & \cellcolor[rgb]{0.9,1.0,0.9}\bm{$1.3417\scriptstyle{\pm0.00020}$} & $0.5138$ \\
        \bottomrule
    \end{tabularx}
    \vspace{-0.4cm}
    \label{table:ablation-su}
\end{table}

\section{Conclusions And Limitations}
In summary, we propose a novel lifelong policy adaptation method named LoopSR based on looping data collection in real and continual training in the simulated reconstruction. The core of LoopSR is a transformer-based encoder that builds up an effective latent space from real-world trajectories for efficient retrieval and decoding. Autoencoder architecture and contrastive loss are implemented for generalizability and robustness. Decent techniques are employed to consolidate a comprehensive pipeline that exhibits extraordinary performance in both simulation and real-world experiments.


There are several limitations to address in future work. Further theoretical analysis is needed to better understand the contributions of the robot and environment to real-world dynamics, as self-cognition is crucial for human-like activities. Additionally, our method excludes visual perception. 
An ambitious future goal is to enable robots to reconstruct the real world in simulations using visual input and to learn new skills directly.

\bibliographystyle{IEEEtran}
\bibliography{IEEEabrv,icra2025}

@article{NFL_supervised,
  title={The lack of a priori distinctions between learning algorithms},
  author={Wolpert, David H},
  journal={Neural computation},
  volume={8},
  number={7},
  pages={1341--1390},
  year={1996},
  publisher={MIT Press One Rogers Street, Cambridge, MA 02142-1209, USA journals-info~…}
}

@article{NFL_optim,
  title={No free lunch theorems for optimization},
  author={Wolpert, David H and Macready, William G},
  journal={IEEE transactions on evolutionary computation},
  volume={1},
  number={1},
  pages={67--82},
  year={1997},
  publisher={IEEE}
}

@inproceedings{rma,
title={Rma: Rapid motor adaptation for legged robots},
author={Kumar, Ashish and Fu, Zipeng and Pathak, Deepak and Malik, Jitendra},
booktitle={Robotics: Science and Systems},
year={2021}
}

@inproceedings{dreamwaq,
  title={DreamWaQ: Learning robust quadrupedal locomotion with implicit terrain imagination via deep reinforcement learning},
  author={Nahrendra, I Made Aswin and Yu, Byeongho and Myung, Hyun},
  booktitle={2023 IEEE International Conference on Robotics and Automation (ICRA)},
  pages={5078--5084},
  year={2023},
  organization={IEEE}
}

@Misc{unitree,
  title = {Unitree Robotics},
  author = {Unitree},
  howpublished = {\url{https://www.unitree.com/}},
  year = {2022},
}

@inproceedings{randomization1,
  title={Domain randomization for transferring deep neural networks from simulation to the real world},
  author={Tobin, Josh and Fong, Rachel and Ray, Alex and Schneider, Jonas and Zaremba, Wojciech and Abbeel, Pieter},
  booktitle={2017 IEEE/RSJ international conference on intelligent robots and systems (IROS)},
  pages={23--30},
  year={2017},
  organization={IEEE}
}

@inproceedings{randomization2,
  title={Sim-to-real transfer of robotic control with dynamics randomization},
  author={Peng, Xue Bin and Andrychowicz, Marcin and Zaremba, Wojciech and Abbeel, Pieter},
  booktitle={2018 IEEE international conference on robotics and automation (ICRA)},
  pages={3803--3810},
  year={2018},
  organization={IEEE}
}

@inproceedings{randomization3,
  title={Dynamics randomization revisited: A case study for quadrupedal locomotion},
  author={Xie, Zhaoming and Da, Xingye and van de Panne, Michiel and Babich, Buck and Garg, Animesh},
  booktitle={2021 IEEE International Conference on Robotics and Automation (ICRA)},
  year={2021},
}

@inproceedings{rudin2022massive,
  title={Learning to walk in minutes using massively parallel deep reinforcement learning},
  author={Rudin, Nikita and Hoeller, David and Reist, Philipp and Hutter, Marco},
  booktitle={Conference on Robot Learning},
  pages={91--100},
  year={2022},
  organization={PMLR}
}

@article{IsaacGym,
  author    = {Viktor Makoviychuk and
               Lukasz Wawrzyniak and
               Yunrong Guo and
               Michelle Lu and
               Kier Storey and
               Miles Macklin and
               David Hoeller and
               Nikita Rudin and
               Arthur Allshire and
               Ankur Handa and
               Gavriel State},
  title     = {Isaac Gym: High Performance GPU-Based Physics Simulation For Robot
               Learning},
  journal   = {CoRR},
  year      = {2021},
}

@inproceedings{mujoco,
  title={Mujoco: A physics engine for model-based control},
  author={Todorov, Emanuel and Erez, Tom and Tassa, Yuval},
  booktitle={2012 IEEE/RSJ international conference on intelligent robots and systems},
  pages={5026--5033},
  year={2012},
  organization={IEEE}
}

@article{PPO,
  author    = {John Schulman and
               Filip Wolski and
               Prafulla Dhariwal and
               Alec Radford and
               Oleg Klimov},
  title     = {Proximal Policy Optimization Algorithms},
  journal   = {CoRR},
  year      = {2017},
}

@article{SAC,
  author    = {Tuomas Haarnoja and
               Aurick Zhou and
               Kristian Hartikainen and
               George Tucker and
               Sehoon Ha and
               Jie Tan and
               Vikash Kumar and
               Henry Zhu and
               Abhishek Gupta and
               Pieter Abbeel and
               Sergey Levine},
  title     = {Soft Actor-Critic Algorithms and Applications},
  journal   = {CoRR},
  year      = {2018},
}

@article{lee2020challenging,
  title={Learning quadrupedal locomotion over challenging terrain},
  author={Lee, Joonho and Hwangbo, Jemin and Wellhausen, Lorenz and Koltun, Vladlen and Hutter, Marco},
  journal={Science robotics},
  volume={5},
  number={47},
  pages={eabc5986},
  year={2020},
  publisher={American Association for the Advancement of Science}
}

@inproceedings{egocentric,
  title={Legged Locomotion in Challenging Terrains using Egocentric Vision},
  author={Agarwal, Ananye and Kumar, Ashish and Malik, Jitendra and Pathak, Deepak},
  booktitle={6th Annual Conference on Robot Learning},
  year={2022}
}

@article{tan2018sim2real,
  title={Sim-to-real: Learning agile locomotion for quadruped robots},
  author={Tan, Jie and Zhang, Tingnan and Coumans, Erwin and Iscen, Atil and Bai, Yunfei and Hafner, Danijar and Bohez, Steven and Vanhoucke, Vincent},
  journal={arXiv preprint arXiv:1804.10332},
  year={2018}
}

@inproceedings{smith2022legged,
  title={Legged robots that keep on learning: Fine-tuning locomotion policies in the real world},
  author={Smith, Laura and Kew, J Chase and Peng, Xue Bin and Ha, Sehoon and Tan, Jie and Levine, Sergey},
  booktitle={2022 International Conference on Robotics and Automation (ICRA)},
  pages={1593--1599},
  year={2022},
  organization={IEEE}
}

@article{smith2022walk,
  title={Demonstrating a walk in the park: Learning to walk in 20 minutes with model-free reinforcement learning},
  author={Smith, Laura and Kostrikov, Ilya and Levine, Sergey},
  journal={Robotics: Science and Systems (RSS) Demo},
  volume={2},
  number={3},
  pages={4},
  year={2023}
}

@inproceedings{smith2024grow,
  title={Grow your limits: Continuous improvement with real-world rl for robotic locomotion},
  author={Smith, Laura and Cao, Yunhao and Levine, Sergey},
  booktitle={2024 IEEE International Conference on Robotics and Automation (ICRA)},
  pages={10829--10836},
  year={2024},
  organization={IEEE}
}

@inproceedings{dreamer,
  title={Dream to Control: Learning Behaviors by Latent Imagination},
  author={Hafner, Danijar and Lillicrap, Timothy and Ba, Jimmy and Norouzi, Mohammad},
  booktitle={International Conference on Learning Representations},
  year={2019}
}

@inproceedings{wu2023daydreamer,
  title={Daydreamer: World models for physical robot learning},
  author={Wu, Philipp and Escontrela, Alejandro and Hafner, Danijar and Abbeel, Pieter and Goldberg, Ken},
  booktitle={Conference on robot learning},
  pages={2226--2240},
  year={2023},
  organization={PMLR}
}

@InProceedings{simsaim,
    author    = {Chen, Xinlei and He, Kaiming},
    title     = {Exploring Simple Siamese Representation Learning},
    booktitle = {Proceedings of the IEEE/CVF Conference on Computer Vision and Pattern Recognition (CVPR)},
    year      = {2021},
}

@misc{infoNCE,
      title={Representation Learning with Contrastive Predictive Coding}, 
      author={Aaron van den Oord and Yazhe Li and Oriol Vinyals},
      year={2019},
      eprint={1807.03748},
      archivePrefix={arXiv},
      primaryClass={cs.LG},
      url={https://arxiv.org/abs/1807.03748}, 
}

@article{supervised_con,
  title={Supervised contrastive learning},
  author={Khosla, Prannay and Teterwak, Piotr and Wang, Chen and Sarna, Aaron and Tian, Yonglong and Isola, Phillip and Maschinot, Aaron and Liu, Ce and Krishnan, Dilip},
  journal={Advances in neural information processing systems},
  volume={33},
  pages={18661--18673},
  year={2020}
}

@article{con_gcrl,
  title={Contrastive learning as goal-conditioned reinforcement learning},
  author={Eysenbach, Benjamin and Zhang, Tianjun and Levine, Sergey and Salakhutdinov, Russ R},
  journal={Advances in Neural Information Processing Systems},
  volume={35},
  pages={35603--35620},
  year={2022}
}

@inproceedings{CURL,
  title={Curl: Contrastive unsupervised representations for reinforcement learning},
  author={Laskin, Michael and Srinivas, Aravind and Abbeel, Pieter},
  booktitle={International conference on machine learning},
  pages={5639--5650},
  year={2020},
  organization={PMLR}
}

@article{R3M,
  title={R3m: A universal visual representation for robot manipulation},
  author={Nair, Suraj and Rajeswaran, Aravind and Kumar, Vikash and Finn, Chelsea and Gupta, Abhinav},
  journal={arXiv preprint arXiv:2203.12601},
  year={2022}
}

@article{reward_repre_gcrl,
  title={Vip: Towards universal visual reward and representation via value-implicit pre-training},
  author={Ma, Yecheng Jason and Sodhani, Shagun and Jayaraman, Dinesh and Bastani, Osbert and Kumar, Vikash and Zhang, Amy},
  journal={arXiv preprint arXiv:2210.00030},
  year={2022}
}

@article{representation_survey,
  title={Representation learning: A review and new perspectives},
  author={Bengio, Yoshua and Courville, Aaron and Vincent, Pascal},
  journal={IEEE transactions on pattern analysis and machine intelligence},
  volume={35},
  number={8},
  pages={1798--1828},
  year={2013},
  publisher={IEEE}
}

@article{repre_multimodal_sur,
  title={Deep multimodal representation learning: A survey},
  author={Guo, Wenzhong and Wang, Jianwen and Wang, Shiping},
  journal={Ieee Access},
  volume={7},
  pages={63373--63394},
  year={2019},
  publisher={IEEE}
}

@article{repre_con_sur,
  title={Contrastive representation learning: A framework and review},
  author={Le-Khac, Phuc H and Healy, Graham and Smeaton, Alan F},
  journal={Ieee Access},
  volume={8},
  pages={193907--193934},
  year={2020},
  publisher={IEEE}
}

@inproceedings{repre_rl,
  title={Decoupling representation learning from reinforcement learning},
  author={Stooke, Adam and Lee, Kimin and Abbeel, Pieter and Laskin, Michael},
  booktitle={International conference on machine learning},
  pages={9870--9879},
  year={2021},
  organization={PMLR}
}

@article{repre_rl2,
  title={Integrating state representation learning into deep reinforcement learning},
  author={De Bruin, Tim and Kober, Jens and Tuyls, Karl and Babu{\v{s}}ka, Robert},
  journal={IEEE Robotics and Automation Letters},
  volume={3},
  number={3},
  pages={1394--1401},
  year={2018},
  publisher={IEEE}
}

@article{VAE,
  title={Auto-encoding variational bayes},
  author={Kingma, Diederik P},
  journal={arXiv preprint arXiv:1312.6114},
  year={2013}
}

@article{transfer_learning_survey,
  title={Transfer learning in deep reinforcement learning: A survey},
  author={Zhu, Zhuangdi and Lin, Kaixiang and Jain, Anil K and Zhou, Jiayu},
  journal={IEEE Transactions on Pattern Analysis and Machine Intelligence},
  year={2023},
  publisher={IEEE}
}

@inproceedings{transfer_learning2,
  title={Policy invariance under reward transformations: Theory and application to reward shaping},
  author={Ng, Andrew Y and Harada, Daishi and Russell, Stuart},
  booktitle={Icml},
  volume={99},
  pages={278--287},
  year={1999}
}

@inproceedings{robot_specific,
  title={Learning modular neural network policies for multi-task and multi-robot transfer},
  author={Devin, Coline and Gupta, Abhishek and Darrell, Trevor and Abbeel, Pieter and Levine, Sergey},
  booktitle={2017 IEEE international conference on robotics and automation (ICRA)},
  pages={2169--2176},
  year={2017},
  organization={IEEE}
}

@article{Hilbert,
  title={Foundation policies with hilbert representations},
  author={Park, Seohong and Kreiman, Tobias and Levine, Sergey},
  journal={arXiv preprint arXiv:2402.15567},
  year={2024}
}

@inproceedings{TAE,
  title={Generalizable Task Representation Learning for Offline Meta-Reinforcement Learning with Data Limitations},
  author={Zhou, Renzhe and Gao, Chen-Xiao and Zhang, Zongzhang and Yu, Yang},
  booktitle={Proceedings of the AAAI Conference on Artificial Intelligence},
  volume={38},
  number={15},
  pages={17132--17140},
  year={2024}
}

@article{Gait_representation,
  title={Learning multiple gaits within latent space for quadruped robots},
  author={Wu, Jinze and Xue, Yufei and Qi, Chenkun},
  journal={arXiv preprint arXiv:2308.03014},
  year={2023}
}

@inproceedings{walk_these_ways,
  title={Walk these ways: Tuning robot control for generalization with multiplicity of behavior},
  author={Margolis, Gabriel B and Agrawal, Pulkit},
  booktitle={Conference on Robot Learning},
  pages={22--31},
  year={2023},
  organization={PMLR}
}

@article{repre_seq,
  title={A recurrent latent variable model for sequential data},
  author={Chung, Junyoung and Kastner, Kyle and Dinh, Laurent and Goel, Kratarth and Courville, Aaron C and Bengio, Yoshua},
  journal={Advances in neural information processing systems},
  volume={28},
  year={2015}
}

@inproceedings{repre_sto_vvp,
  title={Stochastic Variational Video Prediction},
  author={Babaeizadeh, Mohammad and Finn, Chelsea and Erhan, Dumitru and Campbell, Roy H and Levine, Sergey},
  booktitle={International Conference on Learning Representations},
  year={2018}
}

@article{skill-transfer,
  title={Skill transfer and discovery for sim-to-real learning: A representation-based viewpoint},
  author={Ma, Haitong and Ren, Zhaolin and Dai, Bo and Li, Na},
  journal={arXiv preprint arXiv:2404.05051},
  year={2024}
}

@inproceedings{reward_encoding,
  title={Unsupervised Zero-Shot Reinforcement Learning via Functional Reward Encodings},
  author={Frans, Kevin and Park, Seohong and Abbeel, Pieter and Levine, Sergey},
  booktitle={Forty-first International Conference on Machine Learning},
  year={2024}
}

@article{ASE,
   title={ASE: large-scale reusable adversarial skill embeddings for physically simulated characters},
   volume={41},
   ISSN={1557-7368},
   url={http://dx.doi.org/10.1145/3528223.3530110},
   DOI={10.1145/3528223.3530110},
   number={4},
   journal={ACM Transactions on Graphics},
   publisher={Association for Computing Machinery (ACM)},
   author={Peng, Xue Bin and Guo, Yunrong and Halper, Lina and Levine, Sergey and Fidler, Sanja},
   year={2022},
   month=jul, pages={1–17} }

@article{CRL_definition,
  title={A definition of continual reinforcement learning},
  author={Abel, David and Barreto, Andr{\'e} and Van Roy, Benjamin and Precup, Doina and van Hasselt, Hado P and Singh, Satinder},
  journal={Advances in Neural Information Processing Systems},
  volume={36},
  year={2024}
}

@article{CRL_replay,
  title={Continual reinforcement learning with multi-timescale replay},
  author={Kaplanis, Christos and Clopath, Claudia and Shanahan, Murray},
  journal={arXiv preprint arXiv:2004.07530},
  year={2020}
}

@inproceedings{CRL_SER,
  title={Selective experience replay for lifelong learning},
  author={Isele, David and Cosgun, Akansel},
  booktitle={Proceedings of the AAAI Conference on Artificial Intelligence},
  volume={32},
  number={1},
  year={2018}
}

@article{CRL_PR_distill,
  title={Reset \& Distill: A Recipe for Overcoming Negative Transfer in Continual Reinforcement Learning},
  author={Ahn, Hongjoon and Hyeon, Jinu and Oh, Youngmin and Hwang, Bosun and Moon, Taesup},
  journal={arXiv preprint arXiv:2403.05066},
  year={2024}
}

@article{CRL_DET,
  title={Disentangling transfer in continual reinforcement learning},
  author={Wolczyk, Maciej and Zaj{\k{a}}c, Micha{\l} and Pascanu, Razvan and Kuci{\'n}ski, {\L}ukasz and Mi{\l}o{\'s}, Piotr},
  journal={Advances in Neural Information Processing Systems},
  volume={35},
  pages={6304--6317},
  year={2022}
}

@article{AMP1,
  title={Learning robust and agile legged locomotion using adversarial motion priors},
  author={Wu, Jinze and Xin, Guiyang and Qi, Chenkun and Xue, Yufei},
  journal={IEEE Robotics and Automation Letters},
  year={2023},
  publisher={IEEE}
}

@inproceedings{amp2,
  title={Adversarial motion priors make good substitutes for complex reward functions},
  author={Escontrela, Alejandro and Peng, Xue Bin and Yu, Wenhao and Zhang, Tingnan and Iscen, Atil and Goldberg, Ken and Abbeel, Pieter},
  booktitle={2022 IEEE/RSJ International Conference on Intelligent Robots and Systems (IROS)},
  pages={25--32},
  year={2022},
  organization={IEEE}
}

@article{transformer,
  title={Attention is all you need},
  author={Vaswani, A},
  journal={Advances in Neural Information Processing Systems},
  year={2017}
}

@article{DT,
  title={Decision transformer: Reinforcement learning via sequence modeling},
  author={Chen, Lili and Lu, Kevin and Rajeswaran, Aravind and Lee, Kimin and Grover, Aditya and Laskin, Misha and Abbeel, Pieter and Srinivas, Aravind and Mordatch, Igor},
  journal={Advances in neural information processing systems},
  volume={34},
  pages={15084--15097},
  year={2021}
}

@misc{GPT,
  title={Improving language understanding by generative pre-training},
  author={Radford, Alec},
  year={2018},
  journal={OpenAI Research},
  howpublished={\url{https://cdn.openai.com/research-covers/language-unsupervised/language_understanding_paper.pdf}}
}

@inproceedings{
HIB,
title={Bridging the Sim-to-Real Gap from the Information Bottleneck Perspective},
author={Haoran He and Peilin Wu and Chenjia Bai and Hang Lai and Lingxiao Wang and Ling Pan and Xiaolin Hu and Weinan Zhang},
booktitle={8th Annual Conference on Robot Learning},
year={2024},
url={https://openreview.net/forum?id=Bq4XOaU4sV}
}

\addtolength{\textheight}{-12cm}  

\end{document}